%% file: root.tex
\def\year{2018}\relax
\documentclass[letterpaper]{article}
\usepackage{aaai18}
\usepackage{times}
\usepackage{helvet}
\usepackage{courier}
\usepackage{url}
\usepackage{graphicx}
\usepackage{color}
\usepackage{amsmath,amssymb,bbm}
\usepackage{algorithm,algorithmicx}
\usepackage{algpseudocode}
\usepackage{enumerate}

\algnewcommand\algorithmicinput{\textbf{Init:}}
\algnewcommand\Init{\item[\algorithmicinput]}

\begin{document}

\title{Deep TAMER: Interactive Agent Shaping in High-Dimensional State Spaces}

%\author{Paper ID: 2506}
\author{
Garrett Warnell\textsuperscript{1},
Nicholas Waytowich\textsuperscript{1,2},
Vernon Lawhern\textsuperscript{1},
Peter Stone\textsuperscript{3}\\
\textsuperscript{1}{U.S. Army Research Laboratory},
\textsuperscript{2}{Columbia University, New York},
\textsuperscript{3}{The University of Texas at Austin}\\
\texttt{\{garrett.a.warnell.civ,nicholas.r.waytowich.civ\}@mail.mil},\\
\texttt{vernon.j.lawhern.civ@mail.mil},
\texttt{pstone@cs.utexas.edu}
}

\maketitle

\input{abstract}
\input{introduction}
\input{formulation}
\input{method}
\input{experiments}
\input{summary}
\input{acknowledgments}
\input{references}

\end{document}

%% file: abstract.tex
%!TEX root = root.tex
\begin{abstract}
While recent advances in deep reinforcement learning have allowed autonomous learning agents to succeed at a variety of complex tasks, existing algorithms generally require a lot of training data.
One way to increase the speed at which agents are able to learn to perform tasks is by leveraging the input of human trainers.
Although such input can take many forms, real-time, scalar-valued feedback is especially useful in situations where it proves difficult or impossible for humans to provide expert demonstrations.
Previous approaches have shown the usefulness of human input provided in this fashion (e.g., the TAMER framework), but they have thus far not considered high-dimensional state spaces or employed the use of deep learning.
In this paper, we do both: we propose {\em Deep TAMER}, an extension of the TAMER framework that leverages the representational power of deep neural networks in order to learn complex tasks in just a short amount of time with a human trainer.
We demonstrate Deep TAMER's success by using it and just 15 minutes of human-provided feedback to train an agent that performs better than humans on the Atari game of {\sc Bowling} - a task that has proven difficult for even state-of-the-art reinforcement learning methods.
\end{abstract}

%% file: introduction.tex
%!TEX root = root.tex
\section{Introduction}
\label{sec:intro}
Many tasks that we would like autonomous agents to be able to accomplish can be thought of as the agent making a series of decisions over time.
For example, if an autonomous robot is tasked to navigate to a specific goal location in a physical space, it does so by making a series of decisions regarding which particular movement actions it should execute at every instant.
Solutions to these types of {\em sequential decision making} problems can be specified by {\em policies}, i.e., mappings from the agent's state (e.g., its physical location within the environment) to actions that the agent might take (e.g., to move in a certain direction). 
Armed with a particular policy, an agent may decide what actions to take by first estimating its state and then acting according to the output of that policy.

\begin{figure}[ht]
  \begin{center}
    \includegraphics[width=0.47\textwidth]{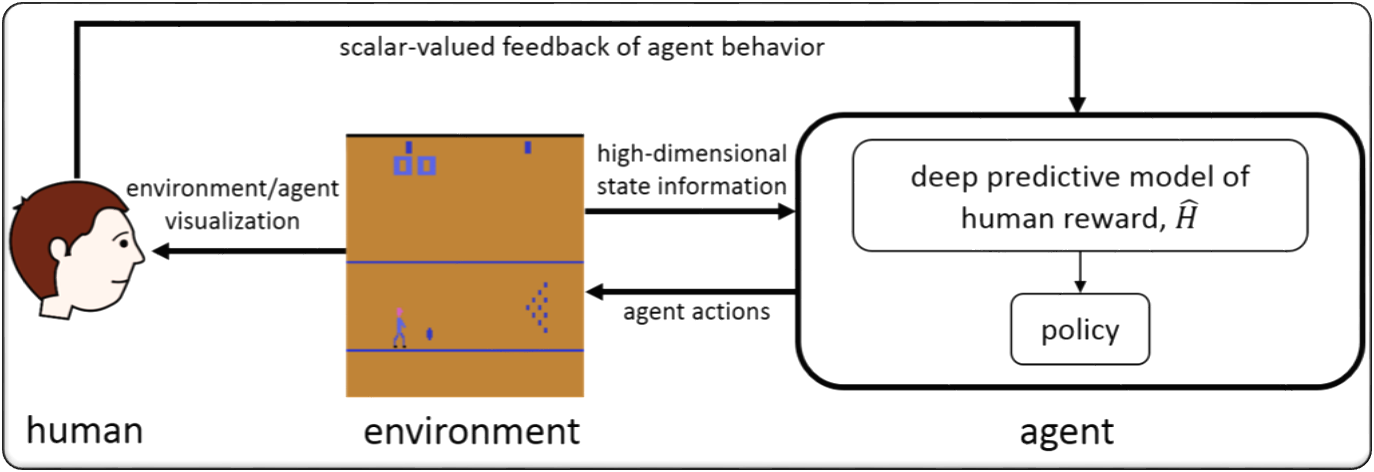}
    \caption{\footnotesize{The Deep TAMER framework proposed in this paper. A human observes an autonomous agent trying to perform a task in a high-dimensional environment and provides scalar-valued feedback as a means by which to shape agent behavior. Through the interaction with the human, the agent learns the parameters of a deep neural network, $\hat{H}$, that is used to predict the human's feedback. This prediction then drives the agent's behavior policy. In this paper, we specifically consider the Atari game of {\sc Bowling}, which uses a pixel-level state space.}}
    \label{fig:paradigm}
  \end{center}
\end{figure}

In some very simple cases, human experts are able to completely specify in advance (i.e., ``hand code") policies that allow agents to accomplish certain tasks.
For many complex tasks, however, specifying policies in such a manner is prohibitively difficult.
Instead, good policies are often found automatically through machine learning methods such as reinforcement learning \cite{Sutton1998}.
These techniques can compute optimal policies without explicit human direction using only data available to the agent as it interacts with its environment.

While most state-of-the-art reinforcement learning approaches to finding optimal decision-making policies enjoy reasonable success, they also usually require a large amount of time and/or data.
This requirement is primarily due to the fact that interesting tasks often exhibit high-dimensional state spaces such as those that arise when the agent's state information includes raw image data.
In the absence of any {\em a priori} information, performing reinforcement learning over these state spaces requires that a large number of parameters be estimated from scratch, and, as a result, even state-of-the-art techniques typically require extremely large amounts of training data, which often translates to large amounts of time before agents are able to learn good policies.
For example, recent methods in {\em deep reinforcement learning} must acquire tens of millions of video frames worth of experience in a simulator before they can learn good strategies for playing Atari games (e.g., \cite{Mnih2015a}).

Here, we are concerned with situations where fast training time is of critical importance.
Like others before us, we propose to achieve more rapid agent learning by adapting the typical reinforcement learning paradigm such that it is able to exploit the availability of a human trainer, and we develop a novel method by which to do so.
Human trainers are useful because they typically possess a good understanding of the task they would like the agent to perform at the outset, and this understanding can be thought of as critical {\em a priori} information that might allow an agent to drastically reduce its training time. 
This observation was recently made and exploited in a series of papers outlining the TAMER framework \cite{Knox2009,Knox2012,Knox2013,Knox2015} (see Figure \ref{fig:paradigm}), in which it was shown that agents learning directly from non-expert human trainers can improve over agent-only learning both in terms of sample complexity and overall performance.
While there has been some follow-on work that examined the quality of human feedback as a function of factors relating to user engagement \cite{Li2013b,Li2015,Li2017}, until now, this framework was only shown to work in low-dimensional state spaces.

In this paper, we specifically seek to augment the TAMER framework such that it may be utilized in higher-dimensional state spaces.
We propose to do so by incorporating recent function approximation techniques from deep learning that have enabled a number of recent successes in reinforcement learning over high-dimensional spaces.
In particular, the contributions of our work are twofold:
\begin{enumerate}[\indent (1)]
	\item We propose specific enhancements to TAMER that enable its success in high-dimensional state spaces in a framework we call Deep TAMER.
	\item We quantify the performance difference between TAMER and the proposed technique in an environment with high-dimensional state features.
\end{enumerate}

In order to evaluate Deep TAMER, we focus here specifically on the Atari game of {\sc Bowling}.
Training agents that are successful at this game has proven difficult for even state-of-the-art deep reinforcement learning methods.
For example, even with training times on the order of days, several recently-proposed frameworks for deep reinforcement learning are only able to train agents that obtain raw scores ranging from $35$ to $70$ out of a maximum of $270$ \cite{Mnih2016}.
In contrast, we show in this paper that, using the proposed technique, human trainers can train better agents within just a few minutes of interactive training.
Moreover, we show that human trainers are often able to train the agent to achieve a better score on {\sc Bowling} than the trainers themselves.

\subsection{Related Work}
While there is a large body of literature on techniques that allow autonomous agents to learn through interaction with human trainers, none of it addresses the problem domain or uses the methodology we propose to study here.

One area of research on learning from human interaction that has received a great deal of attention is focused on the specific problem of {\em learning from demonstration} \cite{Schaal1997,Argall2009,Hussein2017}.
Techniques developed to address this problem typically take as input observations of a human performing a task, and aim to use these demonstrations in order to compute policies that allow the autonomous agent to also accomplish the demonstrated task.
While learning from demonstration techniques often exhibit impressive performance in terms of training time, the necessity of demonstration data may sometimes prove infeasible: an expert human demonstrator may not always be available, or, for difficult tasks, may not even even exist.
This is also problematic for the related area of {\em inverse reinforcement learning} \cite{Ng2000,Abbeel2004}, in which autonomous agents attempt to learn the reward function - and, through that, a useful policy - using explicit demonstrations of the task.
Moreover, since the goal in this setting is often to mimic the human, performance on the underlying task is typically capped at what was exhibited by the demonstrator, which makes difficult the task of training autonomous agents that are capable of super-human performance.

Another area of research that aims to use explicit human interaction to increase the speed at which autonomous agents can learn to perform tasks is that of {\em reward shaping} \cite{Skinner1938,Randløv1998b,Ng1999,Devlin2012}.
Reward shaping techniques do not require demonstrations, but rather that humans modify directly a low-level {\em reward function} that is used to specify the desired task.
Assuming the human has a good understanding of the task and the intricacies of how to appropriately modify the underlying reward function, these approaches can also be quite effective.
However, as with learning from demonstration, it is not often the case that a human has either the level of understanding or the proficiency in machine learning to be able to interact with the agent in this way.

Most closely related to the work presented here, several methods have recently been proposed that allow an agent to learn from interaction with a non-expert human \cite{Thomaz2006,Macglashan2014,Loftin2014,Loftin2016,Peng2016}.
In this work, we consider the TAMER (Training an Agent Manually via Evaluative Reinforcement) technique \cite{Knox2009,Knox2012,Knox2013,Knox2015} for training autonomous agents in particular.
TAMER allows an autonomous agent to learn from non-expert human feedback in real time through a series of {\em critiques}: the human trainer observes the agent's behavior and provides scalar feedback indicating how ``good" or ``bad" they assess the current behavior to be.
The agent, in response, tries to estimate a hidden function that approximates the human's feedback and adjusts its behavior to align with this estimate.
While the TAMER framework has proven successful in several limited tasks, existing work has not yet considered the high-dimensional state spaces often encountered in many more complex tasks of interest.

Separately, very recent work has yielded a major breakthrough in reinforcement learning by utilizing new function approximation techniques from {\em deep learning} \cite{Krizhevsky2012,LeCun2015}.
This advancement has spurred a huge leap forward in the ability of reinforcement learning techniques to learn in high-dimensional state spaces, such as those encountered in even simple Atari games \cite{Mnih2015a,Mnih2016}.
Deep learning has also been applied to several of the areas described above, including learning from demonstration \cite{Hester2017} and inverse reinforcement learning \cite{Wulfmeier2015a}.
However, the effect of using deep learning in techniques that learn from human interaction remains largely understudied and could enable highly efficient human-agent learning for many tasks of interest.

One notable place in which deep learning and learning from human interaction has been studied is in the recent work of \cite{Christiano2017}.
There, deep learning was applied to a paradigm in which the learner actively queries humans to compare behavior examples during the learning process.
While this work is indeed similar to ours, we can highlight two key differences.
First, the method discussed in \cite{Christiano2017} requires the learner to have access to, and use in the background, a simulator during the human interaction, which may not always be possible in real-world situations such as those that might arise in robotics.
The method we describe here does not require access to a simulator during the interaction.
Second, the technique proposed by \cite{Christiano2017} requires on the order of $10$ million learning time steps to be executed in the simulator during interaction with the human, which is only possible with access to extremely powerful hardware.
The method we propose here learns from a human in real-time without a simulator, and requires just a few thousand time steps of interaction to learn a good policy for the environment considered here.
This enables its success with standard computing hardware (e.g., a consumer laptop).

\begin{figure*}[ht]
  \includegraphics[width=\textwidth]{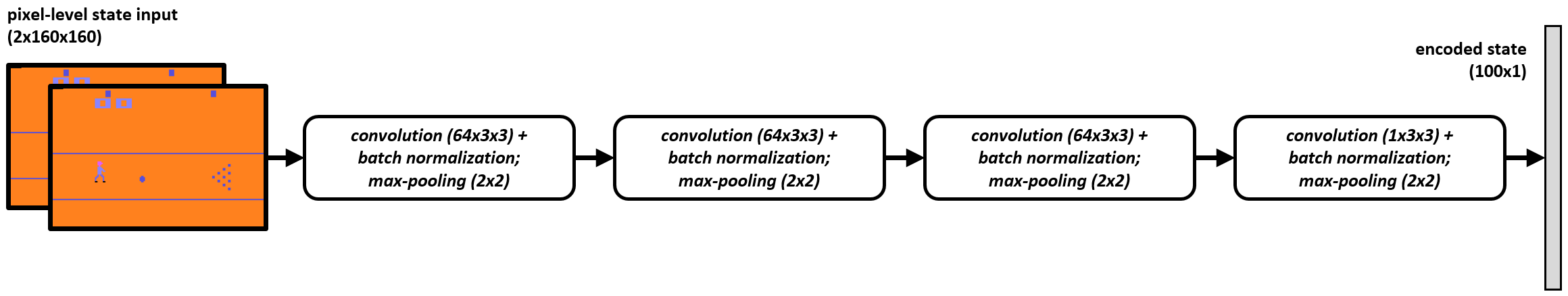}
  \caption{\footnotesize{The specific network structure for the state encoder, $f(\cdot,\boldsymbol{\theta}_f)$, that we use as the fixed front half of $\hat{H}$ in this work. There are $76,035$ parameters to be learned, and the optimal set, $\boldsymbol{\theta}_f^*$, is found during an autoencoder pre-training phase (with a decoder, not shown, that has symmetric structure) over states obtained while the agent executes a random policy in simulation.}}
  \label{fig:autoencoder}
\end{figure*}

%% file: formulation.tex
%!TEX root = root.tex
\section{Problem Formulation}
\label{sec:pf}
In this paper, we are concerned with answering the following specific question: {\em what is the impact of using deep neural networks on the effectiveness of learning from real-time, scalar-valued human feedback in high-dimensional state spaces}?
In this section, we formulate this problem more precisely.
While we adopt a formulation similar to that proposed in \cite{Knox2009}, there will be two main ways in which ours differs: {\em (1)} in the choice of function class the agent uses to represent the feedback coming from the human, and {\em (2)} in the specific optimization techniques used to train the agent.

We consider the agent learning problem in the context of classical sequential decision making \cite{Bellman1957}.
Specifically, let $\mathcal{S}$ denote the set of {\em states} in which an autonomous agent can find itself, and let $\mathcal{A}$ denote the {\em actions} that this agent may execute.
As the agent selects and executes actions $(\mathbf{a}_1,\mathbf{a}_2,\ldots)$ which result in a state trajectory $(\mathbf{s}_0,\mathbf{s}_1,\mathbf{s}_2,\mathbf{s}_3,\ldots)$, we assume here that a human trainer observes the state trajectory and periodically provides scalar-valued feedback signals, $(h_1,h_2,\ldots)$, that convey their assessment of the agent's behavior.
Here, we assume that larger $h$ correspond to a more positive assessment from the human.
Critically, we also assume that the human provides this feedback according to some hidden function $H(\cdot,\cdot): \mathcal{S} \times \mathcal{A} \rightarrow \mathbb{R}$ that the human implicitly understands.
The agent learning problem we consider here is that of computing an estimate of this function, which we denote as $\hat{H}$, such that it is a good match for $H$.
Given the current estimate $\hat{H}$, we assume that the agent behaves myopically according to a fixed action-selection policy $\pi$ that selects the next action as the one that maximizes the predicted human feedback, i.e., $\pi(\mathbf{s}) = \max_\mathbf{a} \hat{H}(\mathbf{s},\mathbf{a})$.

Given this fixed behavior policy, the problem described above is exactly one of supervised machine learning: the agent observes inputs to $H$ in the form of its own experience $\mathbf{x} = (\mathbf{s},\mathbf{a},t^s,t^e)$ and observations related to its output in the form of the human's feedback $\mathbf{y} = (h,t^f)$, where $(t^s,t^e)$ is the interval of time the agent spent in state $\mathbf{s}$ before taking action $\mathbf{a}$ and $t^f$ is the time at which the human's feedback was observed.
From these observations, the agent is tasked with computing the estimate $\hat{H}$.
Importantly, we do not assume that there is a one-to-one mapping between inputs and outputs.
That is, we expect that each observed $\mathbf{y}$ corresponds to several recently-observed $\mathbf{x}$s, and, moreover, that some $\mathbf{x}$ may have no corresponding feedback information at all.
We encode this assumption in the learning process by using the following loss function to judge the quality of $\hat{H}$:
\begin{align}
\ell( \hat{H} \; ; \; \mathbf{x}, \mathbf{y} ) = w(t^s,t^e,t^f)\left[ \hat{H}(\mathbf{s},\mathbf{a}) - h \right]^2 \; ,
\label{eq:weightedsquaredloss}
\end{align}
where $w$ is a scalar-valued weighting function that is larger for $(\mathbf{x},\mathbf{y})$ pairs for which we hypothesize that the human intended $h$ to apply to $(\mathbf{s},\mathbf{a})$. 
The exact form of $w$ will be discussed in the next section, but, importantly, it has the property that it is zero-valued when $t^f$ is less than $t^s$ and becomes negligible when $t^f$ is sufficiently greater than $t^e$.
These events correspond to when the human feedback was provided before the state-action pair occurred, or long after it, respectively.

Using (\ref{eq:weightedsquaredloss}) as our loss function amounts to judging the quality of $\hat{H}$ according to its predictive power in the context of our prior belief about what the trainer intended to evaluate with their feedback.
Therefore, the learning problem becomes one of simply computing an estimate that minimizes (\ref{eq:weightedsquaredloss}) over the set of available observations.
Since we would like the agent to learn throughout the interaction with the human and not after, we define our problem as one of {\em online} supervised learning where we treat observations as realizations of random variables and seek to minimize the loss in a statistical sense.
Here, we accomplish this goal by seeking to find an $\hat{H}$ that minimizes the expected value of the loss, i.e.,
\begin{align}
\hat{H}^* = \arg\min_{\hat{H}} \mathbb{E}_{\mathbf{x},\mathbf{y}}\left[ \ell(\hat{H}\; ; \; \mathbf{x},\mathbf{y}) \right] \;
\label{eq:satisticalmin}
\end{align}
where the expectation is taken with respect to $(\mathbf{x},\mathbf{y})$ pairs generated during the real-time interaction between the human and the agent.
Programs of the form of (\ref{eq:satisticalmin}) are amenable to online solution techniques, and we shall describe ours in the next section.

%% file: method.tex
%!TEX root = root.tex
\section{Method}
\label{sec:meth}
We shall adopt a common technique for approaching problems formulated as in (\ref{eq:satisticalmin}) when observations are only sequentially available: that of {\em stochastic gradient descent} (SGD) \cite{Robbins1951,Bottou1998}.
Broadly speaking, SGD computes incremental estimates $\hat{H}_k$ via a descent procedure that uses instantaneous approximations of the gradient, i.e.,
\begin{align}
\hat{H}_{k+1} = \hat{H}_k - \eta_k \nabla_{\hat{H}} \ell( \hat{H}_k \; ; \; \mathbf{x}_{i_k},\mathbf{y}_{j_k} ) \; .
\label{eq:generalsgd}
\end{align}
Above, $k$ is the iteration index (i.e., the number of times feedback-experience information has been used to update $\hat{H}$) and $\nabla_{\hat{H}} \ell( \hat{H}_k \; ; \; \mathbf{x}_{i_k},\mathbf{y}_{j_k} )$ denotes the gradient of $\ell$ taken with respect to the first argument given an $(\mathbf{x},\mathbf{y})$ pair sampled uniformly at random from the stream of experience observations $(\mathbf{x}_1,\mathbf{x}_2,\ldots)$ and feedback observations $(\mathbf{y}_1,\mathbf{y}_2,\ldots)$.

\subsection{Importance Weights}
\label{subsec:iw}
In selecting (\ref{eq:weightedsquaredloss}) as our loss function, we have opted to use a {\em weighted} squared loss, with weights given by $w$.
We use these weights as {\em importance weights}, i.e., we use them to bias our solution toward those $(\mathbf{x},\mathbf{y})$ pairs for which we hypothesize the human intended their feedback information $\mathbf{y}$ to apply to agent experience $\mathbf{x}$.
Since we select the $k^\text{th}$ $(\mathbf{x},\mathbf{y})$ pair in (\ref{eq:generalsgd}) uniformly at random from the stream of all experience and feedback observations, simply using a loss of $(\hat{H}(\mathbf{s},\mathbf{a}) - h)^2$ would amount to assuming that all human feedback applied to all experience.
This assumption is clearly at odds with the intent of human trainers providing real-time feedback while watching an agent behave, and so, instead, we assume that human trainers intend for their feedback to apply only to {\em recent} agent behavior.
More specifically, we compute importance weights, $w(t^s,t^e,t^f)$, as the probability that feedback provided at time $t^f$ applies to a state-action pair that occurred during the time interval $(t^s,t^e)$ according to an assumed probability distribution $f_{delay}$.
That is,
\begin{align}
w(t^s,t^e,t^f) = \int_{t^f - t^e}^{t^f - t^s} f_{delay}(t) dt \; .
\label{eq:importanceweights}
\end{align}
While there are many distributions one might adopt for $f_{delay}$ (the interested reader should refer to \cite{Knox2009} for further discussion), we shall use the continuous uniform distribution over the interval $[0.2,4]$.
Thus, each observed feedback $\mathbf{y}$ will only have nonzero $w$ for those $\mathbf{x}$ observed between $4$ and $0.2$ seconds before the feedback occurred.
For learning efficiency, when performing SGD (\ref{eq:generalsgd}), we do not sample $(\mathbf{x},\mathbf{y})$ pairs for which the corresponding $w$ is zero since such pairs will result in no update to $\hat{H}$.

\begin{algorithm}
  \label{alg:deeptamer}
  \begin{algorithmic}[1]
  	%\Init Replay memory $D$ to capacity $N$
  	%\Init Credit window buffer $w$ to capacity $L$
    \Require{pre-initialized $\hat{H}_0$, step size $\eta$, buffer update interval $b$}
    \Init $j=0$, $k=0$
    \For{$i = 1,2,\ldots$}
      \State {\bf observe} state $\mathbf{s}_i$
      \State {\bf execute} action $a_{i} = \text{arg\,max}_\mathbf{a} \hat{H}_k(\mathbf{s}_{i},\mathbf{a})$
      \State $\mathbf{x}_i = (\mathbf{s}_i,\mathbf{a}_i,t_i,t_{i+1})$
      \If{new feedback $\mathbf{y} = (h,t^f)$}
        \State $j = j+1$
        \State $\mathbf{y}_j = \mathbf{y}$
        \State $\mathcal{D}_j = \Big\{ (\mathbf{x},\mathbf{y}_j) \; | \; w(\mathbf{x},\mathbf{y}_j \neq 0 \Big\}$
        \State $\mathcal{D} = \mathcal{D} \cup \mathcal{D}_j$
        \State {\bf compute} $\hat{H}_{k+1}$ using SGD update (\ref{eq:generalsgd}) and mini-batch $\mathcal{D}_j$
        \State $k = k+1$
      \EndIf
      \If{ mod($i$,$b$)==0 and $\mathcal{D} \neq \emptyset$}
        \State {\bf compute} $\hat{H}_{k+1}$ using SGD update (\ref{eq:generalsgd}) and mini-batch sampled from $\mathcal{D}$
        \State $k = k+1$
      \EndIf
    \EndFor
  \end{algorithmic}
  \caption{The Deep TAMER algorithm.}
\end{algorithm}

\subsection{Deep Reward Model}
\label{subsec:deepreward}
In order to efficiently learn the human reward function when the dimension of the state space is large, we propose to model this function using a deep neural network.
More specifically, since we are particularly interested in state spaces that are comprised of images (which exhibit dimensionality on the order of tens of thousands), we assume $\hat{H}$ takes the form of a deep convolutional neural network (CNN), $f$, followed by a neural network comprised of several fully-connected layers, $z$, i.e., $\hat{H}(\mathbf{s},\mathbf{a}) = z(f(\mathbf{s}),\mathbf{a})$.
In order to efficiently learn the parameters of this network under the constraints of both limited training time and limited human input, we adopt two strategies: {\em (1)} we pre-train the CNN portion of $\hat{H}$ using an autoencoder, and {\em (2)} we use a {\em feedback replay buffer} as a means by which to increase the rate of learning.

\begin{figure}[t!]
  \includegraphics[width=0.48\textwidth]{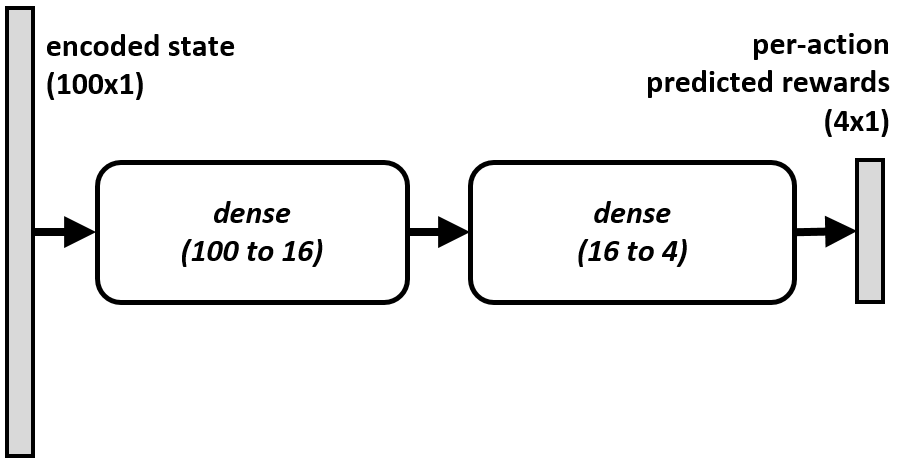}
  \caption{\footnotesize{The specific network structure for the fully-connected portion of $\hat{H}$, $z(f(\mathbf{s}),\mathbf{a})$. The action input specifies which component of the output vector is used as the final output of $\hat{H}$.}}
  \label{fig:tamer_network}
\end{figure}

\subsubsection{Deep Autoencoder}
\label{subsubsec:autoenc}
As a means by which to immediately reduce the number of parameters of $\hat{H}$ that the agent must estimate during real-time interaction with the human, we pre-train the convolutional layers of the network using an autoencoder.
Our autoencoder is comprised of two functions: an {\em encoder}, $f(\cdot; \boldsymbol{\theta}_f)$, and a {\em decoder}, $g(\cdot;\boldsymbol{\theta}_g)$, parameterized by $\boldsymbol{\theta}_f$ and $\boldsymbol{\theta}_g$, respectively.

The encoder we use is a deep CNN that accepts states $\mathbf{s}$ of dimension $d$ and outputs encoded states $f(\mathbf{s})$ of dimension $p << d$.
For the Atari {\sc Bowling} environment, we let $\mathbf{s}$ represent the two most-recent $160 \times 160$ game images, i.e., $d = 51,200$, and we use an encoder structure that produces a $p = 100$-dimensional output.
The exact structure is given in Figure \ref{fig:autoencoder}.

The decoder we use during pre-training is the mirror image of $f$: it is a deep deconvolutional neural network that accepts the encoded states and outputs decoded states $g(f(\mathbf{s}))$ of dimension $d$.
The parameters of the encoder and decoder are jointly found by trying to minimize the reconstruction error over a set of training states.
That is, the optimal values for $\boldsymbol{\theta}_f$ and $\boldsymbol{\theta}_g$ are given by
\begin{equation}
(\boldsymbol{\theta}_f^*, \boldsymbol{\theta}_g^*) = \arg\min_{(\boldsymbol{\theta}_f, \boldsymbol{\theta}_g)} \frac{1}{M} \sum_{i=1}^M \lVert \mathbf{s}_i - g( f(\mathbf{s}; \boldsymbol{\theta}_f); \boldsymbol{\theta}_g) \rVert_2^2 \; .
\end{equation}
We acquire the training states for each environment in off-line simulation using a random policy.
After training is complete, we use the resulting encoder, $f(\cdot;\boldsymbol{\theta}_f^*)$, as a fixed front end for $\hat{H}$.

\subsubsection{Feedback Replay Buffer}
\label{subsubsec:feedbackreplaybuffer}
Even with the CNN parameters fixed at $\boldsymbol{\theta}_f^*$, there are still a large number of parameters of $\hat{H}$ that must be learned, namely those of the fully-connected deep network that comprises the second half of the function.
As a means by which to more quickly learn these parameters during the limited amount of real-time human interaction that the agent receives, we perform SGD updates (\ref{eq:generalsgd}) at a fixed rate that typically exceeds the rate at which humans are typically able to provide feedback to the agent.
We are able to do so by storing all observed human feedback and relevant agent experience in a {\em feedback replay buffer}, $\mathcal{D}$, and sampling repeatedly from this buffer with replacement.
More specifically, $\mathcal{D}$ contains the running history of all human feedback information along with, for each individual feedback signal, the set of experience points that yield nonzero importance weights.
That is, $\mathcal{D} = \Big\{ (\mathbf{x}_i,\mathbf{y}_j) \; | \; w(\mathbf{x}_i,\mathbf{y}_j) \neq 0 \Big\}$, where we use the shorthand $w(\mathbf{x},\mathbf{y})$ for the quantity specified in (\ref{eq:importanceweights}).
Because the set of states $\{ \mathbf{x} \; | \; w(\mathbf{x},\mathbf{y}) \neq 0 \}$ all occur within a fixed window of time {\em before} the corresponding $\mathbf{y}$ is observed, this set can be easily obtained and stored immediately upon observation of each $\mathbf{y}$.
In this work, we do not consider limiting the size of $\mathcal{D}$, and simply store all observed human feedback information provided during the interaction (typically on the order of $1,000$ feedback signals during the $15$-minute training sessions that we shall describe in the next section).

Though we continually perform SGD updates at a fixed rate by sampling from $\mathcal{D}$, we also wish to ensure that each human feedback has an immediate effect on the agent's behavior.
Therefore, we perform SGD updates to $\hat{H}$:
\begin{enumerate}[\indent(a)]
  \item whenever the human provides new feedback, using the new feedback information as the data sample; and
  \item at a fixed rate, using data sampled from $\mathcal{D}$,
\end{enumerate}
where the fixed rate is a parameter of the algorithm.
In our experiments, we select this parameter such that it results in buffer updates every $10$ time steps, while we have observed that human feedback is typically provided at a rate of approximately one signal every $25$ time steps (i.e., in $100$ time steps, there are $10$ buffer updates and typically around $4$ updates triggered by new human feedback).
In practice, we perform mini-batch updates using the average gradient computed over several $(\mathbf{x},\mathbf{y})$ samples instead of just one.
In particular, our mini-batches are formed by first sampling several $\mathbf{y}$ and, for each individual $\mathbf{y}$, adding all $(\mathbf{x},\mathbf{y})$ pairs for which the corresponding $w$ is nonzero to the mini-batch.

\subsection{Deep TAMER}
\label{subsec:deeptamer}
While we have already discussed the deep encoder portion of $\hat{H}$, we have not yet discussed the remainder of the network.
For $z$ in $\hat{H}(\mathbf{s},\mathbf{a}) = z(f(\mathbf{s}),\mathbf{a})$, we use a two-layer, fully-connected neural network with $16$ hidden units per layer and one output node per available action, which is similar in input-output structure to value networks used in recent deep reinforcement learning literature \cite{Mnih2015a}. 
The exact structure of $z$ is shown in Figure \ref{fig:tamer_network}.
The overall predicted human reward value for a given state-action pair, $\hat{H}(\mathbf{s},\mathbf{a})$, is found by using $f(\mathbf{s};\boldsymbol{\theta}_f^*)$ as the input to the fully-connected network and selecting the output node corresponding to action $\mathbf{a}$.
During training, errors are only fed back through the single relevant output node.

We term using the deep reward model and the training method described above - including pre-training the autoencoder and using the importance-weighted stochastic optimization technique with the feedback replay buffer - as {\em Deep TAMER}, and propose do so in autonomous agents that aim to learn from real-time human interaction in environments with high-dimensional state spaces.
The complete procedure is summarized in Algorithm 1. %\ref{alg:deeptamer}.

Beyond the use of a deep neural network function approximation scheme, Deep TAMER differs from TAMER in several important ways.
First, the specific loss function used, (\ref{eq:weightedsquaredloss}), is different than the one used by TAMER.
In using (\ref{eq:weightedsquaredloss}), Deep TAMER seeks to minimize a weighted difference between the human reward and the predicted value for each state-action pair {\emph{individually}}.
This is in contrast to the loss function used in the TAMER framework, which is defined for an entire window of samples, i.e., $\ell(\hat{H} \; ; \; \{\mathbf{x}\}_j ,\mathbf{y} ) = \frac{1}{2}\left( h - \sum_j w(\mathbf{x}_j,\mathbf{y})\hat{H}(\mathbf{s}_j,\mathbf{a}_j)^2 \right)$.
We hypothesize that (\ref{eq:weightedsquaredloss}) more faithfully reflects the intuition that the human's reward signal applies to individual state-action pairs.

Another major difference between Deep TAMER and TAMER is in the frequency of learning.
TAMER learns once from each state-action pair, whereas Deep TAMER can learn from each multiple times due to the feedback replay buffer.
Moreover, unlike the replay buffers used in recent deep RL literature, Deep TAMER's feedback replay buffer is used explicitly to address sparse feedback as opposed to overcoming instability during learning.

%% file: experiments.tex
%!TEX root = root.tex
\section{Experiments}
\label{sec:exp}

\begin{figure}[!t]
  \begin{center}
    \includegraphics[scale=0.49]{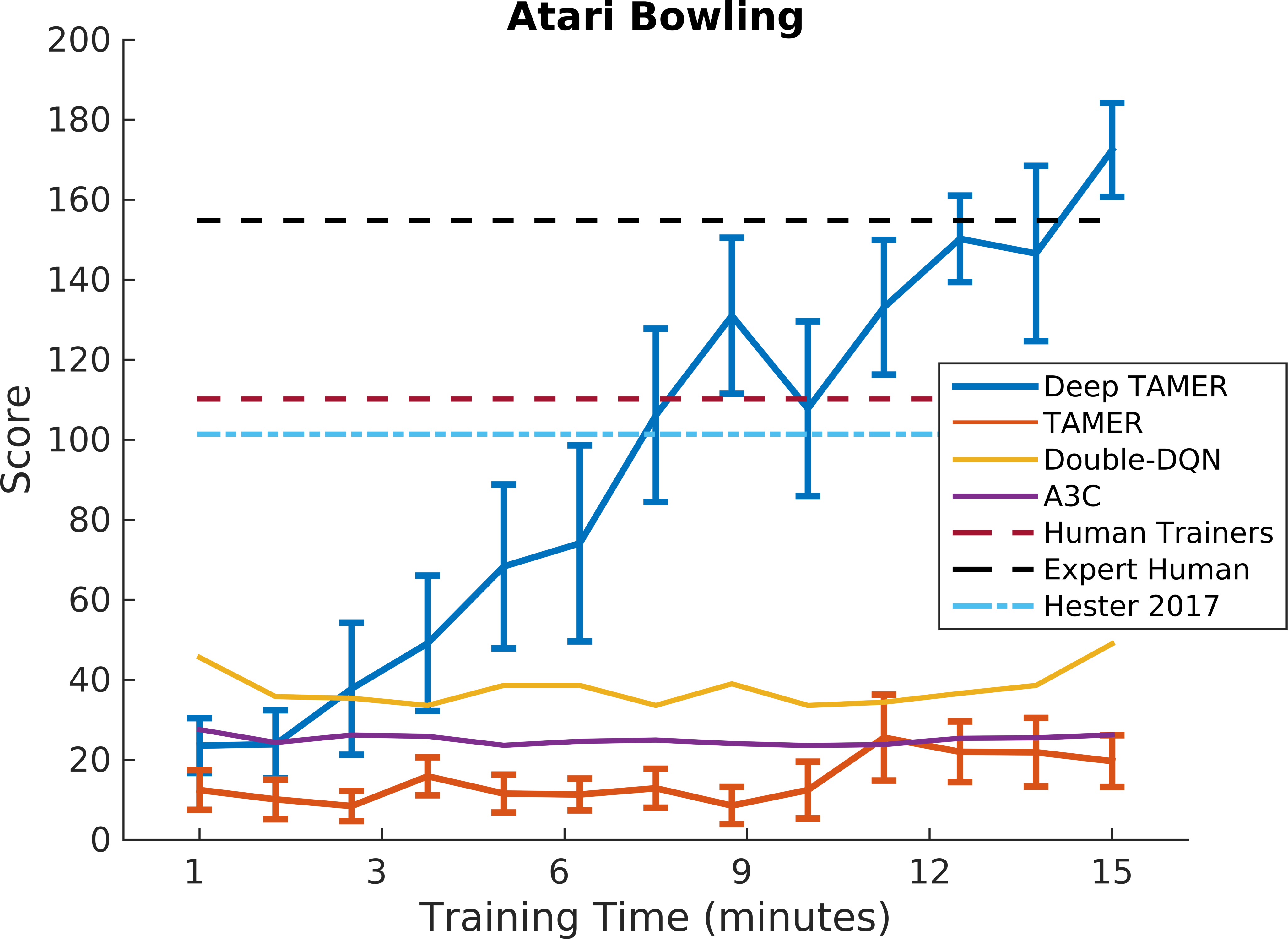}
    \caption{Average score per episode vs wall-clock time for the Atari {\sc Bowling} game. The blue and orange lines show the Deep TAMER and TAMER learning curves averaged over 9 human trainers. The yellow line shows the Double DQN performance. The purple line shows the average A3C performance over all 16 parallel workers. The red dashed line shows the average game score achieved from the human trainers. The black dashed line shows the average game score achieved by the professional human play tester in \cite{Mnih2015a}. The blue dashed line shows the Deep Q-learning from human demonstration method from \cite{Hester2017} which is the previous best method for Atari \textsc{Bowling}.}
    \label{fig:main_comparison}
  \end{center}
\end{figure}

We experimentally evaluated Deep TAMER in the context of the Atari game of {\sc Bowling}, which is an environment that has proven difficult for several recent state-of-the art deep reinforcement learning algorithms.
For our experiments, we used the implementation provided by the Aracade Learning Environment \cite{Bellemare2013} included as part of the OpenAI Gym suite \cite{Brockman2016}.
In this section, we shall describe the environment, detail the procedure by which we had humans train the agent, present the ways in which we evaluated Deep TAMER, and then and discuss the results of our evaluation.

In short, we found that, using Deep TAMER, human trainers were able to train successful {\sc Bowling} agents in just $15$ minutes.
Moreover, we found that agents trained using Deep TAMER outperformed agents trained using state-of-the art deep reinforcement learning techniques as well as agents trained using the original TAMER method proposed in \cite{Knox2009}.
In most cases, the Deep TAMER agents even performed better than the human trainers themselves.
These results demonstrate that the proposed enhancements to TAMER, namely the deep reward model, allow the framework to be successful in environments with high-dimensional state spaces.
Additionally, the favorable comparisons to both deep reinforcement learning techniques and human players indicates the continued utility in the paradigm of autonomous agent learning through real-time human interaction.

\subsection{Atari \textsc{Bowling}}
\label{subsec:atarigames}
A screenshot of the Atari game of {\sc Bowling} is depicted in Figure \ref{fig:paradigm}.
At each time step (we set the rate to approximately $20$ frames per second), an image of the game screen is displayed and the player may select one of four actions: \texttt{no-action}, \texttt{up}, \texttt{down}, and \texttt{bowl}.
We convert the game image to grayscale and use as the game state the $160 \times 160 \times 2$ tensor corresponding to the two most-recent game screens.
Bowling a single ball begins with the player moving the character avatar vertically (using \texttt{up} and \texttt{down}) to a desired position from which to release the ball.
Then, the player must execute the \texttt{bowl} action to start the ball rolling toward the pins.
Finally, the player has the chance to, at any single point during the ball's journey toward the end of the lane, cause the ball to start spinning in the up or down direction, by executing the \texttt{up} or \texttt{down} actions, respectively.
The game proceeds, and is scored, just like common ten-pin bowling, though no extra balls are allowed after the tenth frame, resulting in a maximum score of $270$.

\begin{figure*}[ht]
	\includegraphics[width=\textwidth]{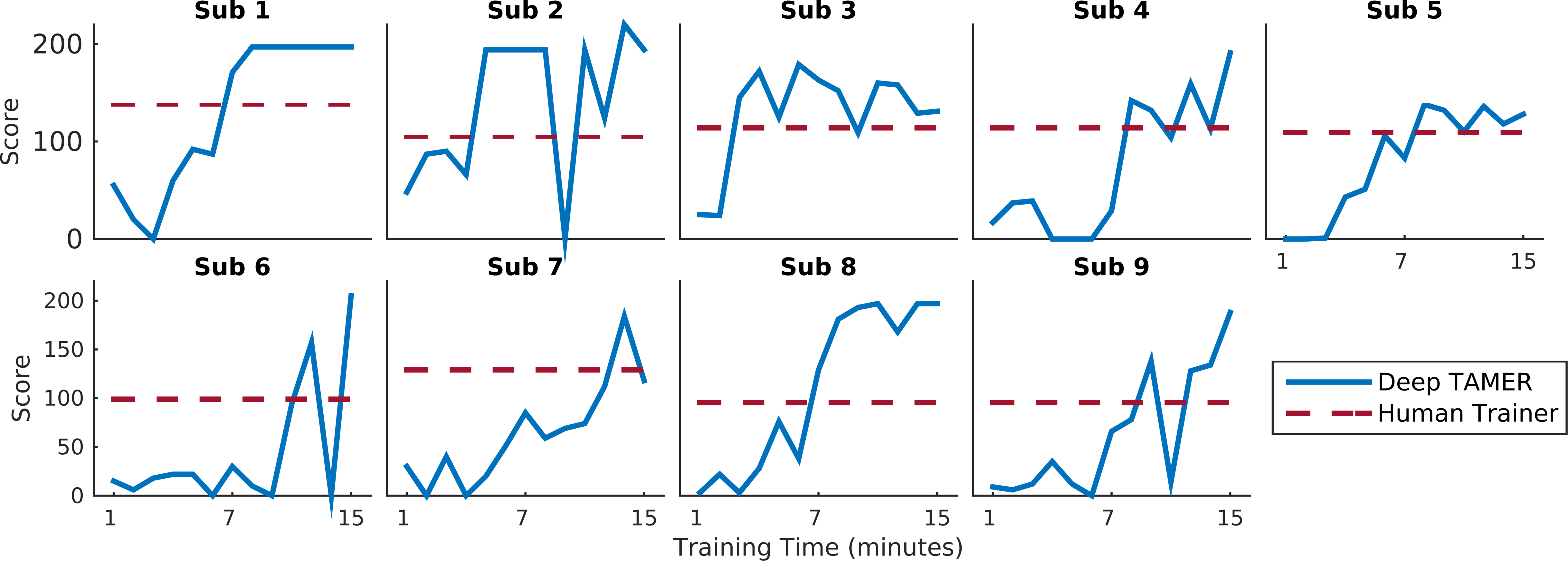}
	\caption{Individual human trainer performance. Each panel shows the Deep TAMER learning curve of game score vs training time for each of the $9$ human trainers in the experiment (solid blue line). The red dashed line shows the average game score achieved by the trainers playing two complete bowling games. All Deep TAMER agents were able to meet or exceed their human trainers after $15$ minutes of training.}
	\label{fig:individual_scores}
\end{figure*}

\subsection{Training Procedure}
\label{subsec:experimentalproc}
To evaluate our system, we had $9$ human trainers train an agent to play {\sc Bowling} using Deep TAMER.
All training was performed using an experimental computer, and trainers were supervised by experimenters.
As a means by which to both provide familiarization with the game and to characterize individual human performance, each trainer first played two ten-frame games each during which we recorded the game score.
Then, each trainer was allowed a $10$-minute training session that was not recorded as a means by which to practice giving feedback to the agent.
During the practice session, trainers were allowed the opportunity to ask the experimenter any questions that they might have.
Finally, after the practice session, the experimenter reset the agent, and the trainer again used Deep TAMER to train an agent for $15$ minutes while the interaction information and other data relating to the agent was recorded.

For comparison,we additionally had $9$ human trainers train an agent to play {\sc Bowling} using TAMER \cite{Knox2009} over the same state space using the same procedure as as above.

\subsection{Evaluation}
We compared the performance of Deep TAMER on Atari {\sc Bowling} against Double-DQN \cite{Hasselt2015}, A3C \cite{Mnih2016}, TAMER \cite{Knox2009}, and the performance of the human trainers themselves.
We used implementations of D-DQN and A3C that have been made available from OpenAI \cite{baselines,baselines2}.
For TAMER, we implemented Algorithm 2 of \cite{Knox2009} using the same credit assignment scheme, with the only difference being that, similar to our $\hat{H}$, a separate linear parameter vector was learned for each discrete action.
Figure \ref{fig:main_comparison} shows the average game score as a function of training time (wall-clock time) for Deep TAMER, TAMER, DDQN and A3C (solid colored lines).
The Deep TAMER and TAMER results were computed by averaging results across $9$ human-trained policies each.
For further comparison, the scores of the human trainers tested in this experiment as well as the expert human game tester reported in \cite{Mnih2015a} are shown as horizontal dashed lines.
Additionally, Deep TAMER was compared to the previously best method on Atari \textsc{Bowling} from \cite{Hester2017} that used human demonstration data to train a deep Q Network (shown as a horizontal dashed line in Figure \ref{fig:main_comparison}).

As expected, the Double-DQN and A3C agents fail to learn a useful policy in the same $15$ minutes that was allotted to the human trainers using Deep TAMER.
Further, even given the full amount of training data required by these methods (i.e., $50$-$100$ million training steps), these algorithms still fail to learn a successful policy for {\sc Bowling} \cite{Mnih2016}, probably due to the sparse reward signal from the environment.
It can be seen that the original TAMER algorithm also failed to learn a useful policy in the allotted time.
This is likely due to the fact that TAMER uses a linear model for the human reward function, which is insufficient for the pixel-based state input that we use here.
Deep TAMER also performs better than the learning from demonstration method from \cite{Hester2017}, indicating an advantage of a reward shaping methods over a demonstration methods when the task is difficult for a human to perform but not difficult for a human to critique.
Finally, as seen in Figure \ref{fig:main_comparison}, Deep TAMER is able to exceed the performance of the human trainers after only $7$ minutes of training, and surpasses the performance of the expert human after $15$ minutes. 

This last point is especially interesting to us as it implies that TAMER is not only useful in that it allows non-experts to train agents that perform well on complex tasks, but that it is also a promising methodology for easily creating agents capable of {\em super-human} performance.
Figure \ref{fig:individual_scores} demonstrates this point in more detail: the Deep TAMER performance for the $9$ human trainers along with their individual play performance is shown.
While performance increase is noisy - likely as a result of the stochastic optimization technique - in all cases, the human trainers were able to train the Deep TAMER agent to play {\sc Bowling} in just $15$ minutes, and in the majority of cases ($6$ out of $9$), they were able train Deep TAMER agents to play better than themselves.

\subsection{Importance Weighting Study}
\begin{figure}[!t]
  \begin{center}
    \includegraphics[scale=0.49]{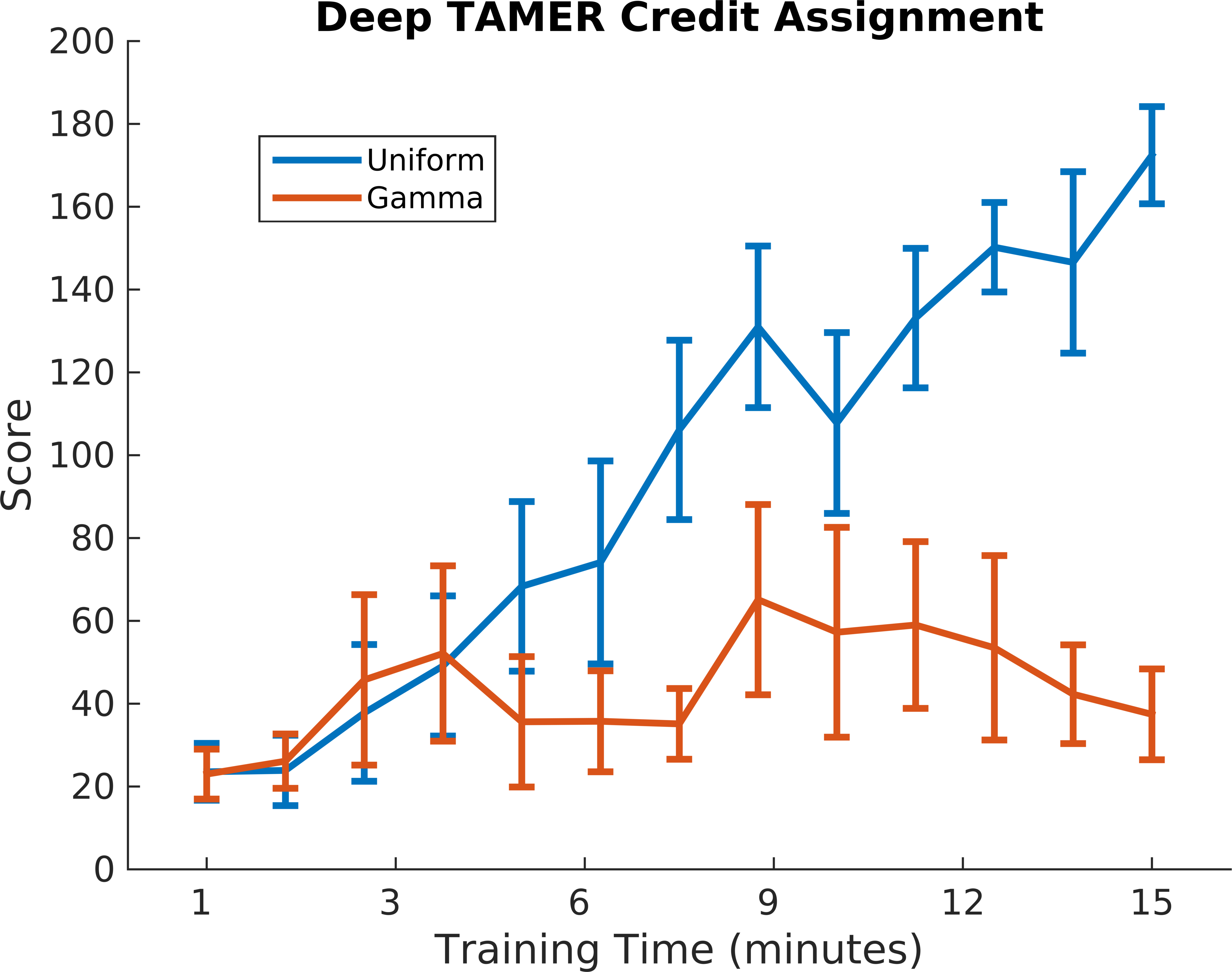}
    \caption{Comparison of different credit assignment distributions for Deep TAMER. The blue curve shows the average over 9 subjects collected using a Uniform $[0.28,4.0]$ distribution and the red curve shows the average of 8 subjects collected using a Gamma($2.0,0.28$) distribution for the credit assignment.}
    \label{fig:credit_comparison}
  \end{center}
\end{figure}

We also performed a small study comparing the effect of using another reasonable importance-weighting scheme: that which results when using a Gamma($2.0,0.28$) distribution as $f_{delay}$ as suggested in \cite{Knox2009}.
A comparison of the training performance under this weighting scheme with that of the one based on the Uniform$[0.28,4.0]$ distribution used throughout this work is shown in Figure \ref{fig:credit_comparison}.
It can be seen that the uniform distribution yielded better results, likely due to the fact that trainers tended to wait until they observed the number of pins that fell before providing feedback.
Since we found that the time between the relevant throw and/or spin actions was typically separated from the time of feedback by between $2$ and $4$ seconds, the uniform scheme resulted in higher importance weights for these critical state-action pairs than those provided by the Gamma distribution.

Due to the difficulty of obtaining large amounts of human interaction data, we did not perform any additional hyperparameter search.
That is, with the exception of the alternative importance-weighting scheme we presented above, we fixed our choices for hyperparameters such as the number of input frames, encoder structure, etc. based on intuition, and we presented here an evaluation of only those choices.
For this same reason, we limited our analysis to the Bowling domain.
There are, of course, a large number of domains and hyperparameter settings that make sense to experiment with, and we hope to do so in future work.

%% file: summary.tex
%!TEX root = root.tex
\section{Summary}
\label{sec:sum}
In this paper, we proposed an extension of the TAMER framework for learning from real-time human interaction.
Our technique, called Deep TAMER, enables the TAMER paradigm to be successful even in environments with high-dimensional state spaces.
This success is due to the use of a deep neural network function model to approximate the human trainer's reward function, and also to the modified supervised learning procedure we proposed to find the parameters of this new model.

We evaluated Deep TAMER on the challenging Atari game of {\sc Bowling} with pixel-level state features, and found that agents trained by humans using Deep TAMER significantly outperformed agents trained by humans using TAMER.

Additionally, our results reaffirm the attractiveness of the overall TAMER paradigm.
After just $15$ minutes of real-time interaction with a human, Deep TAMER agents were able to achieve higher scores than agents trained using state-of-the-art deep reinforcement learning techniques and orders of magnitude more training data.
Moreover, human trainers were actually able to produce agents that actually exceeded their own performance.

%% file: acknowledgments.tex
\section{Acknowledgments}
A portion of this work has taken place in the Learning Agents Research Group (LARG) at UT Austin.  LARG research is supported in part by NSF (IIS-1637736, IIS-1651089, IIS-1724157), Intel, Raytheon, and Lockheed Martin.  Peter Stone serves on the Board of Directors of Cogitai, Inc.   The terms of this arrangement have been reviewed and approved by the University of Texas at Austin in accordance with its policy on objectivity in research.

%% file: references.tex
\bibliographystyle{aaai}
\bibliography{gaw}